\documentclass{article}

\usepackage{PRIMEarxiv}

\usepackage[utf8]{inputenc} 
\usepackage[T1]{fontenc}    
\usepackage[linesnumbered,lined,vlined,ruled]{algorithm2e}
\usepackage{amssymb}
\usepackage{amsmath}
\usepackage{latexsym}
\usepackage{caption}
\usepackage{subcaption}
\usepackage{hyperref}       
\usepackage{url}            
\usepackage{booktabs}       
\usepackage{makecell}
\usepackage{amsfonts}       
\usepackage{nicefrac}       
\usepackage{microtype}      
\usepackage{fancyhdr}       
\usepackage{graphicx}       
\usepackage{textcomp}
\usepackage{multirow}
\usepackage{enumitem}
\usepackage{float}
\usepackage{authblk}

\usepackage{algorithmic}

\title{MemeGraphs: Linking Memes to Knowledge Graphs
}

\author[1,2]{\fontsize{11}{13}\selectfont Vasiliki Kougia}
\author[1]{Simon Fetzel}
\author[1]{Thomas Kirchmair}
\author[1]{Erion Çano}
\author[3]{Sina Moayed Baharlou}
\author[4]{Sahand Sharifzadeh}
\author[1]{Benjamin Roth}

\affil[1]{\fontsize{10}{12}\selectfont University of Vienna, Faculty of computer science, Vienna, Austria}

\affil[ ]{\texttt{\{firstname.lastname,simonf93\}@univie.ac.at}}

\affil[2]{\fontsize{10}{12}\selectfont UniVie Doctoral School Computer Science, Vienna, Austria}

\affil[3]{Boston University, Department of Electrical and Computer Engineering, Boston, MA, USA} \affil[ ]{\texttt{baharlou@bu.edu}}

\affil[4]{Ludwig Maximilians University of Munich, Faculty of Computer Science, Munich, Germany}
\affil[ ]{\texttt{sahand.sharifzadeh@gmail.com}}

\begin{document}
\maketitle

\begin{abstract}
Memes are a popular form of communicating trends and ideas in social media and on the internet in general, combining the modalities of images and text. They can express humor and sarcasm but can also have offensive content. Analyzing and classifying memes automatically is challenging since their interpretation relies on the understanding of 
visual elements, language, and background knowledge. Thus, it is important to meaningfully represent these sources and the interaction between them in order to classify a meme as a whole. In this work, we propose to use scene graphs, that express images in terms of objects and their visual relations, and knowledge graphs as structured representations for meme classification with a Transformer-based architecture. We compare our approach with ImgBERT, a multimodal model that uses only learned (instead of structured) representations of the meme, and observe consistent improvements. We further provide a dataset with human graph annotations that we compare to automatically generated graphs and entity linking. Analysis shows that automatic methods link more entities than human annotators and that automatically generated graphs are better suited for hatefulness classification in memes.
\end{abstract}


\section{Introduction}

Internet memes are items such as images, videos, or twitter posts that are widely shared on social media and typically relate to several subjects, such as politics, social, news, and current internet trends.\footnote{Disclaimer: This paper contains examples of hateful content.} Memes are a popular form of communication and they are often used as a means to express an opinion or stance in a humorous or sarcastic manner, but they can also be hateful and promote problematic content that is likely to hurt specific groups of people and hence be harmful to society in general \cite{Pedro2022}. Thus, analyzing trending memes can provide insight into people's reactions and opinions to important societal matters, as well as to the traits of different groups. This can help in tasks like filtering out harmful memes from internet platforms or extracting user opinions for socioeconomic studies. This work focuses on memes in the form of images with some form of superimposed text (sometimes referred to with the technical term \emph{image macros}) with the goal of detecting hateful content.

\begin{figure*}[t]
\centering
\includegraphics[width=0.9\textwidth]{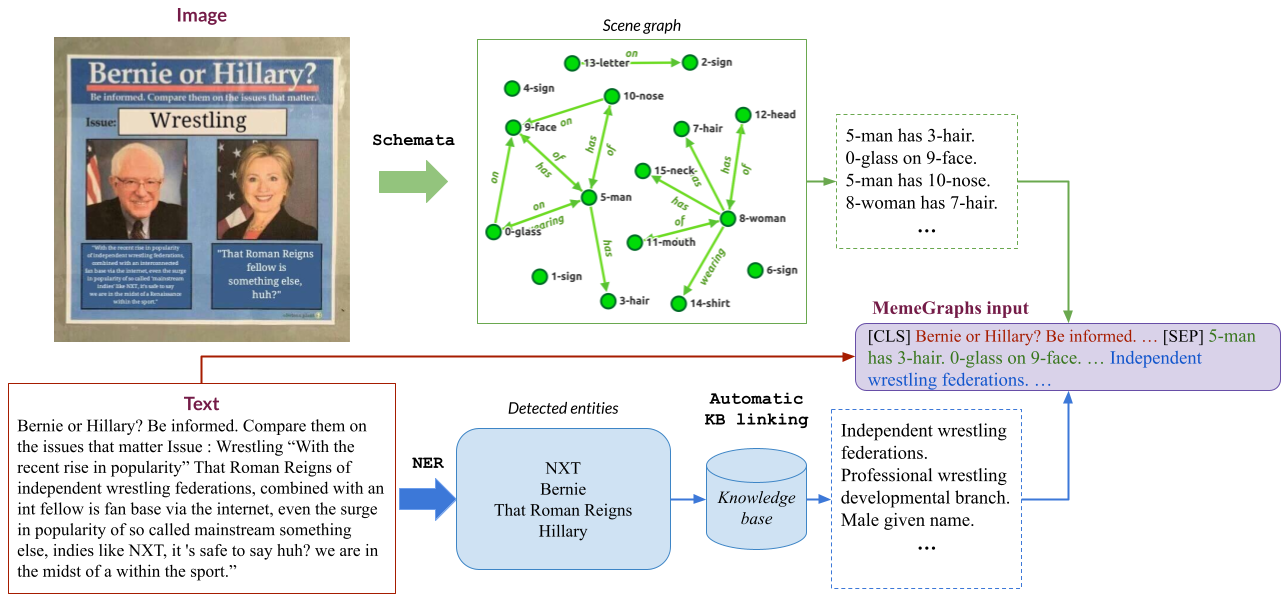}
\caption{The steps performed by the MemeGraphs method. The automatic augmentation consists of scene graphs generated automatically by a pre-trained model (Schemata \cite{Sharifzadeh2021}) and entities detected in the text by a pre-trained Named Entity Recognition (NER) model. Background knowledge for each entity is retrieved from a knowledge base (Wikidata). The final MemeGraphs input is created by concatenating these augmentations and adding them after the [SEP] token following the text of the meme in order to feed it to a Transformer for text classification.}
\label{fig:annotation}
\end{figure*}

Lately, there has been increased interest in deep learning models for analyzing memes and classifying them \cite{Pramanick2021momenta,Dimitrov2021,Fersini2022}, for example in the context of the Hateful Memes competition organized by Facebook \cite{Kiela2020} and the shared task in the Workshop on Online Abuse and Harms (WOAH) 2021 that was the continuation of the first competition \cite{Mathias2021}. Another shared task in Semeval 2022 aimed at detecting misogyny in memes \cite{Fersini2022}. Alongside these shared tasks, the corresponding datasets were published with human annotations capturing important properties such as hatefulness and misogyny. Recent works use multimodal representation learning based on image features from Convolutional Neural Networks (CNNs), Transformer-based language models for the text \cite{Kiela2020,Behera2020,Kougia2021} or multimodal (vision+language) Transformer models in order to classify memes \cite{Kiela2020}.
Some works additionally incorporate specifically extracted image features such as race \cite{Zhu2020}, person attributes \cite{Aggarwal2021,Pramanick2021momenta}, and automatic image captioning output \cite{Das2020,Blaier2021}. 
However, none of these works have included the relations between objects in the form of scene graphs or background knowledge in the form of a text description for the objects depicted in the image.

In this paper, we address the issue of classifying hateful memes by performing an automated augmentation in order to represent the visual information and background knowledge (Fig. \ref{fig:annotation}). We built on the MultiOFF dataset, which contains memes extracted from social media during the 2016 U.S. Presidential Election \cite{MultiOFF}.
An off-the-shelf scene graph generation model was employed to produce scene graphs, which contain detected visual objects and relations between them for each meme, and a NER model to detect entities that we linked to background knowledge. The scene graphs were serialized as text resulting in a unified way to represent all the modalities expressed in a meme. This allows using a (unimodal) text classifier to classify the multimodal memes. Hence, we incorporated the automatically produced augmentations to classify the memes using a text-based Transformer model and show that they can improve its performance. The explicit representation of the image as serialized tokens also provides a more interpretable intermediate representation (compared to hidden layers in multimodal models such as ImgBERT \cite{Kiela2020,Kougia2021}).

Furthermore, in order to examine how the results of the automatic augmentation would deviate from human ones, we performed manual augmentation. Two human evaluators corrected the automatically produced scene graphs and manually added background knowledge. We compare our automatic MemeGraphs method with models operating on the manually augmented data (only for training or for both training and inference) and find that automatic augmentations assist in achieving better results than manual ones.

\paragraph{\textbf{Contribution}} Our contributions can be summarized as follows: 
\begin{itemize}
    \item We propose MemeGraphs, a novel method for classifying memes utilizing scene graphs augmented with knowledge and providing insights for processing multimodal documents.
    \item We show that adding this kind of knowledge to a text-based Transformer model can improve its classification performance. Furthermore, we show that this yields improvements compared to a simple model only using learned representations to classify the memes, such as ImgBERT.
    \item We conduct extensive experiments with manual and automatic settings for obtaining this knowledge and show that the automatic setting of our MemeGraphs method provides more meaningful information.
\end{itemize}

In the following, we discuss related work and present our MemeGraphs method. Subsequently, in Section \ref{sec:benchmark}, we describe the models we implemented and report their results. Finally, we provide a qualitative analysis, including a discussion of the findings of human augmentation, and compare this with the automatic MemeGraphs method.\footnote{The code and data for the MemeGraphs method are available on:\\ \url{https://github.com/vasilikikou/memegraphs}.}

\section{Related Work}
\label{sec:related_work}

Combining text and image inputs is crucial for many tasks, e.g. for image search or (visual) question answering, relying on an image and its caption. Research has shown that adding images to text-based tasks (e.g., machine translation) improves the performance of the models \cite{Yin2020}. However, the meaningful interpretation of text and image and, in particular, the relations between them, still remains challenging \cite{Li2021,Chen2021}. Commonly used approaches rely on Transformer models that are pre-trained on image+text pairs \cite{visualbert,vilbert,uniter,Gan2020,vlbert,oscar}. A step towards better scene understanding is to generate scene graphs \cite{Krishna2017}. 
Scene graphs provide structured knowledge about an image, e.g., objects, relations, and attributes. 
Recent works have shown that we can improve scene graph generation using message propagation between entities \cite{zellers2018neural,yang2018graph}, and by employing background knowledge in form of knowledge graphs \cite{Sharifzadeh2021}, texts \cite{Sharifzadeh2022}, or using feedback connections. In \cite{Sharifzadeh2021}, the authors proposed Schemata, a scene graph generation model consisting of two parts: the backbone module and the relational reasoning component. Additionally, Schemata uses feedback connections to further encourage the propagation of higher-level, class-based knowledge to each neighbor. The backbone is pre-trained on ImageNet~\cite{ImageNet} and the whole network is fine-tuned on Visual Genome~\cite{Krishna2017} on the scene graph classification task.
Scene graphs can help to achieve state-of-the-art results in several visual tasks \cite{Yang2019,Mozes2021}. Inspired by these approaches, we generate scene graphs to represent the visual information contained in memes by using the Schemata model (see Section \ref{sec:memegraphs}).

A specific instance of vision and language tasks is memes classification. To address the need for automatic means that can detect hateful content in memes, datasets and models \cite{MultiOFF,Behera2020,Pramanick2021momenta} were published in the last couple of years, and shared tasks \cite{Kiela2020,Mathias2021,Fersini2022} were organized to attract interest on this task. Methods that have been implemented for hateful memes detection can be grouped into three categories: 1. Unimodal methods that use either only the text or the image as input, 2. Multimodal approaches, where image embeddings from an image encoder are fed to a text model and both models are trained separately, and 3. Multimodal methods, consisting of vision+language Transformers, being pre-trained in a multimodal fashion. Current methods experiment with models from all three categories and focus on improving models from the third category by adding extra features \cite{Behera2020,Pramanick2021momenta,Blaier2021,Lee2021}. These features can be visual attributes extracted from CNNs \cite{Kiela2020,Lee2021,Aggarwal2021} (e.g. objects, entities or demographics), representations from CLIP \cite{Radford2021,Pramanick2021momenta}, automatically generated captions \cite{Das2020,Blaier2021}, etc.

The Hateful Memes Challenge, hosted in 2020 by Facebook, was a binary classification task of hate detection \cite{Kiela2020}.\footnote{\url{https://www.drivendata.org/competitions/64/hateful-memes/}} 
Kiela et al. \cite{Kiela2020} created a dataset with 10,000 memes to which they added counterfactual examples in order to make the task more challenging for unimodal approaches. They experimented with several different settings and found that multimodal methods worked best. An extended version of the Hateful Memes Challenge was included as a shared task in the Workshop on Online Abuse and Harms (WOAH) \cite{Mathias2021}. The same dataset was used but it now included new fine-grained labels for two categories: protected category and attack type. In this shared task, multimodal approaches were dominant as well. A multimodal method introduced by \cite{Kiela2020} and subsequently also used for the shared task in WOAH \cite{Kougia2021} incorporated image embeddings as inputs to a text classifier.\footnote{The method was called Concat BERT in \cite{Kiela2020} and ImgBERT in \cite{Kougia2021}. Here we call it ImgBERT because we use their implementation.} This method belongs to the second category and is an early fusion approach meaning that the image embedding and the text embedding are concatenated before feeding them to the classifier. Different types of image and text components are employed in different works. Specifically, in ImgBERT \cite{Kougia2021}, first, they feed the memes images to a convolutional neural network (CNN) and
extract their embeddings. Then, they provide the text of the meme as input to BERT \cite{Devlin2019} and extract the [CLS] token representation. They concatenate the [CLS] token representation with the embedding of the meme’s image and use the result as input to the classifier. During training only
the text-based BERT part of ImgBERT is trained, while the image embeddings remain frozen.

Another dataset for detecting hateful content in memes is the MultiOFF dataset \cite{MultiOFF}. It contains memes that were extracted from social media during the 2016 U.S. presidential elections. The dataset was first shared on Kaggle and consisted of the image URL for each meme, its text and metadata, e.g., timestamp, author, likes, etc.\footnote{\url{https://www.kaggle.com/datasets/SIZZLE/2016electionmemes}} The authors obtained the images from the URLs and discarded any metadata. In total, this dataset contains 743 memes, which were annotated as hateful or non-hateful. In 
\cite{MultiOFF} the authors experimented with unimodal (text only) and multimodal (text and image) approaches, and the model with the highest F1 score was a CNN operating only on the text of the memes.

The above mentioned datasets focus on hate and offensive speech, but there are also datasets that cover other aspects of harmful content in memes. In \cite{Dimitrov2021}, the authors focused on detecting propaganda in memes. They created and released a dataset with 950 memes extracted from Facebook groups, annotated for 22 different propaganda techniques. In their experiments they used existing unimodal and multimodal models and found that the latter, especially multimodally pre-trained Transformers perform best in their setting.
Recently, a challenge called Multimedia Automatic Misogyny Identification (MAMI) focused on detecting misogyny and its exact form, i.e., stereotype, shaming, objectification and violence in memes \cite{Fersini2022}. In \cite{Pramanick2021momenta}, they studied harm in memes and proposed a framework to detect harmful memes and the entities targeted. The authors also released their dataset with 7,096 memes in total about politics and COVID-19.

The existing challenge sets, resources and models show the importance of analyzing internet memes and the challenge to combine all the modalities that form a meme. However, current works focus on incorporating visual information in the form of individual features like the ones described above or automatically generated captions. We propose a novel approach to represent the visual content of memes using scene graphs, hence "translating" them into text form. Furthermore, current methods only extract entities from the images, but not from the captions or texts. We argue that often this is not sufficient (or feasible), since memes can also incorporate screenshots of text, as it is the case in the MultiOFF dataset. Hence, we approach this problem by extracting the entities from the text in order to obtain more information. We further retrieve background knowledge for each extracted entity, and show that this approach is worthwhile to explore, since it allows for a more grounded and comprehensive automatic interpretation of memes.

\section{MemeGraphs}
\label{sec:memegraphs}

In this section, we describe our method for automatic augmentation of a memes dataset, an approach we call MemeGraphs. Our proposed method consists of three steps: 1. Scene graph construction, 2. Knowledge linking to detected entities and 3. The construction of the final MemeGraphs input. The first two steps are performed automatically by using off-the-shelf models. Hence, the result of MemeGraphs are knowledge graphs representing the meme as a whole, which can be used to classify them, e.g., for hate detection. We build on the MultiOFF dataset, which contains memes extracted from social media during the 2016 U.S. Presidential Election \cite{MultiOFF}. In what follows, we present the three individual steps of MemeGraphs in detail and how the final result is constructed.

\subsection{Scene graphs}

Scene graphs provide information for an image in the form of a graph $SG = \{V, E\}$, where $V$ are the objects depicted in the image (nodes) and $E$ the relations between them (edges). Models that generate scene graphs output a set of relation triplets in the form of $\{object_1, relation, object_2\}$, which constitute the graph.
In our task, we generate scene graphs for memes. Each meme $m$ consists of two modalities $(I_m, T_m) \in I \times T$, where $I$ is the set of images and $T$ is the set of texts of the memes. We employ a scene graph generation model $G$, which takes as input a meme image and outputs a scene graph $SG$ as:
\begin{equation}
SG_m = G(I_m)
\end{equation}

Towards this end, we apply Schemata, a pre-trained scene graph generation model \cite{Sharifzadeh2021} that was trained with a multi-task learning strategy to automatically predict scene graphs. We do not further fine-tune the model on the memes present in our data and we only consider the objects ranking amongst the top 16 objects for the scene graph according to the detection scores.\footnote{This number was chosen after observation of the scene graphs resulting from the memes in order to avoid having an overcrowded scene graph with multiple objects.} Hence, the scene graph generated for each meme is a set of triplets as follows:
\begin{equation}
\label{eq:scene_graphs}
SG_m \subseteq \{(object_i, relation, object_j)\} \textrm{,}
\end{equation}
where $0 \leq i,j \leq 15$, $relation \in R$ and $R$ is the set of all possible relations.

\begin{figure*}[t]
\centering
\includegraphics[width=\textwidth]{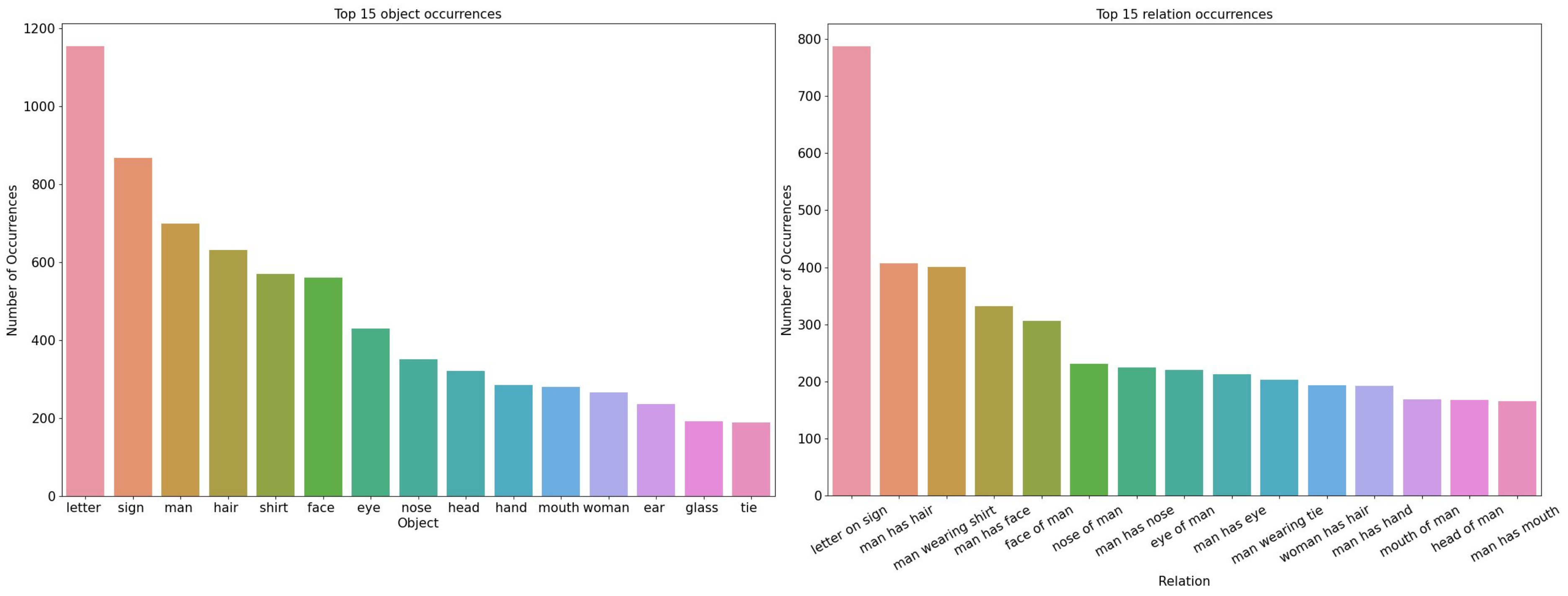}
\caption{The occurrences of the top 15 mostly detected objects (left) and relations (right) based on the automatic annotation.}
\label{fig:frequencies}
\end{figure*}

For 67 memes, Schemata did not produce any output. For the remaining 676 memes, 10,426 relations and 9,666 objects were detected in total. The number of unique objects and relations was 142 and 970 respectively. And while the number of objects detected in an image ranged from 2 to 16, the number of detected relations ranged from 1 to 40. The top 15 most frequently detected object and relation types are shown in Fig. \ref{fig:frequencies}.

\subsection{Knowledge linking}
\label{ssec:linking}

The second step of our MemeGraphs method is to obtain background knowledge for a given meme. In order to do that, first, we employ an NER model. This model detects named entities in a text, which can be person names, organizations, locations, etc. - depending on the task. It can provide useful information that assists to natural language understanding. Here, we feed the text of the meme as input to the NER model and get a list of extracted entities $E$ as $E = NER(T_m)$, where $T_m$ is the text of the meme. Each entity is then searched for in a knowledge base and its related information is retrieved. This way we obtain a text $T_i$ for each entity $E_i$, so for a given meme we have:
\begin{equation}
\label{eq:knowledge}
    KN_m = \{T_i\} = \{KB(E_i)\} 
\end{equation}

For the NER model we employ an off-the-shelf pre-trained Transformer model from spacy.\footnote{\url{https://spacy.io/universe/project/spacy-transformers}} Then, we automatically search for each entity in Wikidata using the API, from which we obtain the description of a data entry for the corresponding entity.\footnote{\url{https://www.wikidata.org/wiki/Wikidata:Main_Page}} Each data entry contains also other information besides the description that could be useful, e.g., the translation of the word in other languages, references to other databases, relevant images and relations e.g., ``instance-of'', ``part-of'', etc.

\subsection{Knowledge graph input}
\label{ssec:kg_input}

The final result of the MemeGraphs augmentation is a serialized text-only representation of the meme that combines the information for all its modalities in one text. 
For the scene graphs we concatenate all the triplets that are detected in a meme (Eq. \ref{eq:scene_graphs}) and thus end up with a text $T_{sg,m}$. For the retrieved background knowledge we concatenate the texts of the $KN$ for each meme (Eq. \ref{eq:knowledge}) and get a text $T_{kn,m}$. The final MemeGraphs input is the text of the meme followed by a [SEP] token and the concatenation of $T_{sg}$ and $T_{kn}$.\footnote{All the texts are concatenated with a full stop.} This text can then be used as input to a classifier that will decide about the hatefulness of the meme. The complete process of the MemeGraphs method is depicted in Algorithm \ref{alg:methodology}.

\begin{algorithm}[H]
\caption{Outline of the MemeGraphs method}
\label{alg:methodology}
\SetAlgoVlined
    \KwData{a set of memes $\mathcal{M}$ consisting of images I and texts T.}
    \KwResult{a set $T_{kg}$ of texts representing the knowledge graphs of the memes.}
    \text{// define a list to save the scene graphs for each meme}\;
    $SG=\{\}$ \;
    \text{// apply the Schemata model}\;
    \For{$m \in \mathcal{M}$} {
        $SG_{m} = G(I_m)$ \;
    }
    \text{// define a list to save the entities of each meme text}\;
    $E=\{\}$ \;
    \text{// apply the NER model}\;
    \For{$m \in \mathcal{M}$} {
        $E_{m} = NER(T_m)$ \;
    }
    \text{// retrieve information from the knowledge base for each entity}\;
    $KN=\{\}$ \;
    \For{$m \in \mathcal{M}$} {
        \For{$e \in E_m$} {
            $KN_m = KB(e)$ \;
        }
    }
    \text{// define lists to save the concatenated texts for each meme}\;
    $T_{sg}, T_{kn}, T_{kg}=\{\}, \{\}, \{\}$ \;
    \For{$m \in \mathcal{M}$} {
        $T_{sg,m} = concat(SG_m)$ \;
        $T_{kn,m} = concat(KN_m)$ \;
        $T_{kg,m} = concat(T_{sg,m}, T_{kn,m})$ \;
    }
    \KwRet $T_{kg}$
\end{algorithm}

\section{Benchmarking}
\label{sec:benchmark}

\subsection{Models}
\label{ssec:benchmark_models}

\begin{figure*}[t]
\centering
\includegraphics[width=0.9\textwidth]{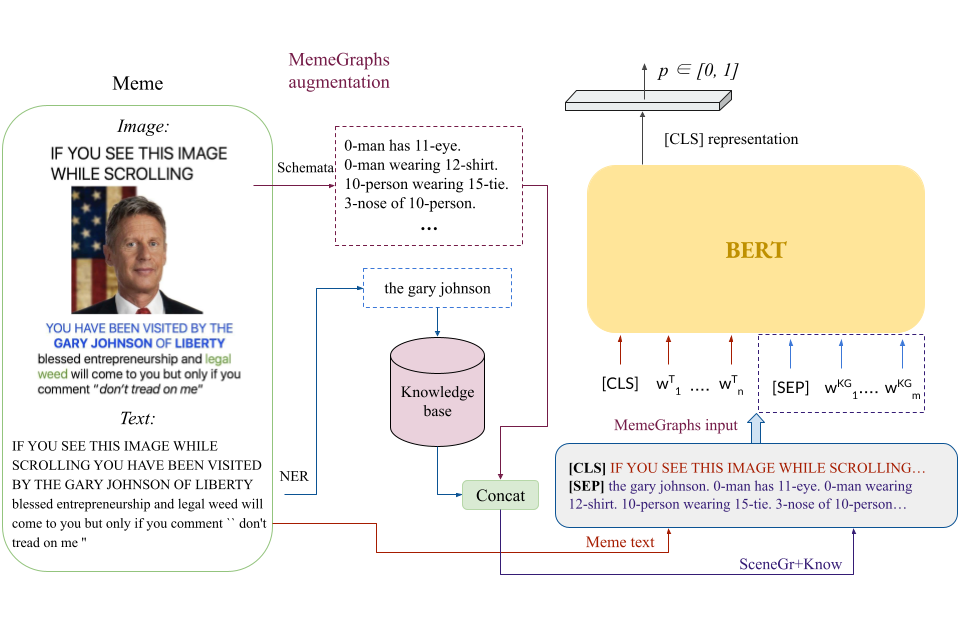}
\caption{The architecture of MemeGraphs[SceneGr+Know]. The scene graph produced by the Schemata model and the background knowledge for each entity are concatenated and given as input to BERT after the text of the meme and the [SEP] token.}
\label{fig:txtbert_kg}
\end{figure*}

To evaluate the results of our automatic augmentation and how it can affect the performance of hateful memes classification, we employed a text-based Transformer with and without the MemeGraphs information as well as a multimodal model. Previous work has shown that using only the text of the memes for classification gives results highly competitive with multimodal methods \cite{Kougia2021,Kiela2020,MultiOFF}. On the other hand, unimodal image-based models have lowest results. We use a pre-trained BERT \cite{Devlin2019} model that takes as input the text of each meme (TxtBERT) \cite{Kougia2021} and different variants of this approach that include the MemeGraphs input. In order to feed the graphs as input to the model, we represent them as a text sequence (see Section \ref{sec:memegraphs}), which is given as an extra text input after the [SEP] token (Fig. \ref{fig:txtbert_kg}). In the model that we call MemeGraphs[SceneGr], this sequence contained only the scene graphs, which were represented by triplets of the detected objects and the relations between them (Eq. \ref{eq:scene_graphs}), e.g., ``0-man has 11-eye. 0-man wearing 12-shirt.''. In the model called MemeGraphs[Know], the text descriptions from the knowledge base corresponding to the detected entities are added as extra input (Eq. \ref{eq:knowledge}). For example, the second input sequence for the meme shown in Figure \ref{fig:txtbert_kg} will be ``American politician, businessman, and 29th Governor of New Mexico.''. While in the MemeGraphs[SceneGr+Know] model, information from the whole knowledge graphs is added, comprising the scene graphs and the background knowledge concatenated with a full stop into one sequence (see Subsection \ref{ssec:kg_input}). In all the models, the [CLS] token is fed into a final linear layer with a sigmoid activation function that produced the probability of the meme being hateful.

In order to compare our MemeGraphs approach with a method that employs learned visual representations, we use ImgBERT, an early fusion multimodal model (see Section \ref{sec:related_work}).

\subsection{Experimental Setup}
\label{ssec:setup}

Each meme was labeled in the original dataset as offensive or non-offensive. The classification was based on the offensiveness labels that exist in the MultiOFF dataset. The dataset is slightly imbalanced with 42\% of the samples being offensive. For training and testing the models, we used the split provided by the authors of the dataset. The split consisted of training, validation, and test sets, which contained 445, 149, and 149 memes respectively. We employed the pre-trained BERT model provided by Hugging Face and fine-tuned it on our dataset.\footnote{\url{https://huggingface.co/docs/transformers/model_doc/bert\#transformers.BertForSequenceClassification}} To obtain image embeddings, we experimented with several pre-trained CNN models and based on the best results we chose DenseNet with 161 layers. The embeddings were extracted from the last pooling layer. The max length was set as the average token length of the training texts, depending on the input of each model. The models were trained using batch size 16, the weighted binary cross entropy loss, and AdamW optimizer with an initial learning rate of 2e-5. We used early stopping based on the validation loss with a patience of 3 epochs. We trained all models 20 times with different seeds and the training lasted between 4 and 6 epochs in each run. During inference, a threshold was used to determine if a meme is hateful or not based on the produced probability. This threshold was set to 0.5 for all the models.

\subsection{Benchmarking results}

For each model, we obtained predictions for the test set from all twenty differently initialized runs. We evaluated each prediction set by calculating the F1 score defining the minority class (offensive class) as positive. In Table \ref{tab:mean_auto_results}, we report the average F1 score over the twenty runs for each method and the corresponding standard error. In Table \ref{tab:best_dev_auto_results}, the test score of the model that achieved the best score on the development set is shown. We observe that MemeGraphs[SceneGr+Know] outperforms the other models when predicting with the best checkpoint (Table \ref{tab:best_dev_auto_results}). On the other hand, when looking at the average, the model with only the background knowledge as input achieves the best F1 score (MemeGraphs[Know]). We generally observe that the methods that incorporate our automatic augmentation (MemeGraphs) outperform the simple text-based fine-tuned BERT. Also, ImgBERT is outperformed showing that when the visual information is employed in the form of scene graphs the performance is improved, compared to when using only the image embeddings.

\begin{table}[t]
\setlength{\tabcolsep}{6pt}
\centering
    \begin{tabular}{l c c c} \hline
    \toprule
      \textbf{Model} & \textbf{P} & \textbf{R} & \textbf{F1} \\ \hline
      ImgBERT & 0.394 \textpm 0.041 & 0.522 \textpm 0.080  & 0.408 \textpm 0.048  \\ \hline \hline
      TxtBERT & \textbf{0.457 \textpm 0.020} & 0.516 \textpm 0.063 & 0.442 \textpm 0.032 \\
      MemeGraphs[SceneGr] & 0.456 \textpm 0.010 & 0.577 \textpm 0.050 & 0.482 \textpm 0.019 \\ 
      MemeGraphs[Know] & 0.446 \textpm 0.011 & 0.574 \textpm 0.045 & \textbf{0.484 \textpm 0.019} \\
      MemeGraphs[SceneGr+Know] & 0.451 \textpm 0.010 & \textbf{0.583 \textpm 0.063} & 0.469 \textpm 0.035 \\ \hline
\end{tabular}
\caption{Test set precision (P), recall (R), and F1 scores averaged over twenty runs with different initializations and standard error of mean.}
\label{tab:mean_auto_results}
\end{table}

\begin{table}[t]
\setlength{\tabcolsep}{6pt}
\centering
    \begin{tabular}{l c c c} \hline
    \toprule
      \textbf{Model} & \textbf{P} & \textbf{R} & \textbf{F1} \\ \hline
      ImgBERT & 0.389 & \textbf{1.000} & 0.560 \\ \hline \hline
      TxtBERT & 0.403  & 0.897 & 0.556 \\
      MemeGraphs[SceneGr] & 0.398 & 0.845 & 0.541 \\
      MemeGraphs[Know] & 0.396 & 0.690 & 0.503 \\
      MemeGraphs[SceneGr+Know] & \textbf{0.426} & 0.948 & \textbf{0.588} \\ \hline
\end{tabular}
\caption{Test set precision (P), recall (R), and F1 scores from the model selected as best on the development set.}
\label{tab:best_dev_auto_results}
\end{table}

\section{Analysis} 
\label{sec:analysis}

\subsection{Human augmentation} 
\label{ssec:humanannotation}

Since the MemeGraphs augmentation is produced automatically it can potentially result in inaccuracies that can deteriorate the performance of the classification models. In order to examine this scenario and also the scene graphs themselves, which were produced by an off-the-shelf model, we performed human augmentation. The augmentation was conducted by two male students, who were doing a Master of Arts (M.A.) in Digital Humanities and the process lasted about 10 weeks. Several discussions took place before the augmentation started for the evaluators to get familiar with this study and understand its scope. They also carefully studied the guidelines before and during the process.

The goal of the human augmentation was to correct the automatically generated scene graphs and add background knowledge by linking the detected objects to a knowledge base (see Fig. \ref{fig:annotation}). For each detected object or relation, the first step for the evaluators was to evaluate if it is correct or not and in case of an incorrect object to correct it. The second step was to link each object to its entry in Wikidata. For example, the detected object ``man'' was linked to the entry for ``man''. The objects represent generic types, e.g., ``man'', ``woman'' etc., but in some cases, a specific instance of the object type might be shown, which we call entity, e.g., the woman depicted is Hillary Clinton (as shown in Fig. \ref{fig:annotation}). Then, the evaluators searched for the entry of each detected object and its entity (if existing) in Wikidata and added the corresponding links. The evaluators worked towards achieving high precision and in order to limit the scope of the augmentation, no new objects or relations were added.

\begin{figure*}[t]
\centering
\includegraphics[width=0.9\textwidth]{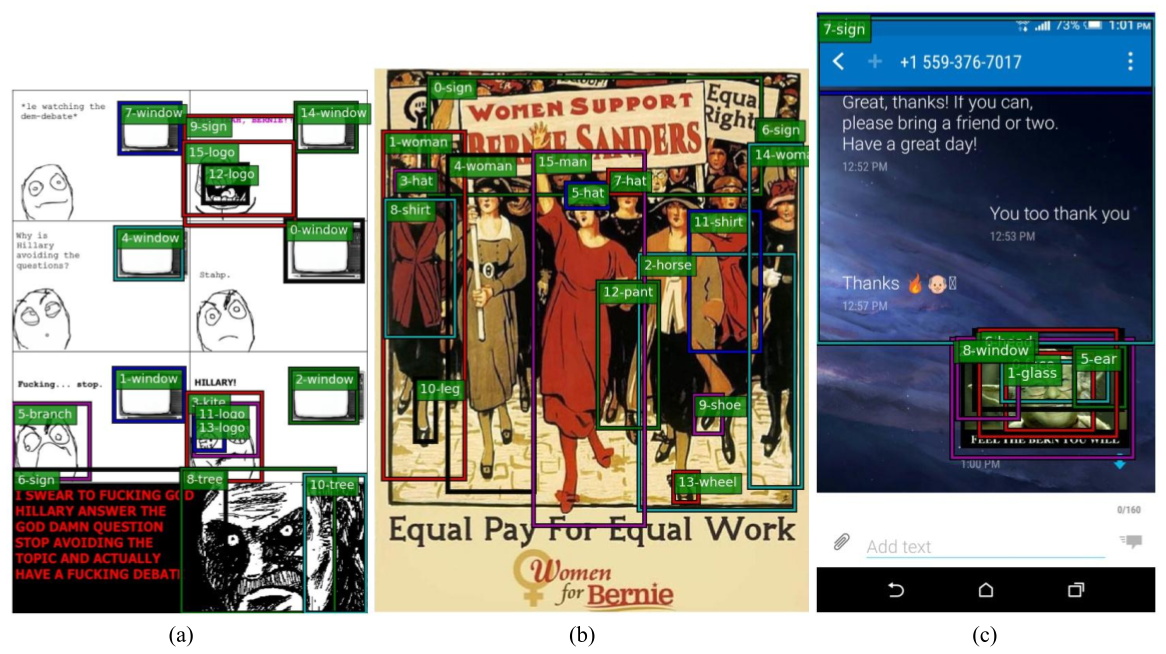}
\caption{Examples of memes after the automatic augmentation. In sub-figures (a) and (b), we see cases where the detected object is incorrect (e.g., ``10-tree'' in (a) and ``13-wheel'' in (b)), but the correct object is not clear to define. In sub-figure (c) we see that the same object, i.e., ``0-face'' can depict two entities.}
\label{fig:annotation_examples}
\end{figure*}

After a first round of augmenting a small sub-sample (around 10\%), preliminary guidelines were created. These guidelines described the process mentioned in the previous paragraph in simple and brief steps. A few systematic difficulties for the Schemata algorithm were encountered in this first round that can be summarized as follows:
\begin{itemize}
    \item The text of the meme appearing on the image was often detected as a ``sign'' and some parts of the text as ``letter''. However, this text is not actually a sign and is not part of the image.
    \item In some cases, the same object was detected multiple times.
    \item Some memes in the dataset were screenshots of text, so there was not any useful visual information extracted from them.
    \item In a few cases, there was a specific entity depicted in the meme, but the corresponding object type was not detected.
\end{itemize}

The above-mentioned difficulties were discussed with the evaluators and the guidelines were revised to include clear instructions for those cases. Based on that and other observations of the evaluators about the data, the following final guidelines were defined:
\begin{itemize}
    \item \textbf{Meme text or sign.} A distinction between a meme's text and an actual sign in the image, e.g., a sticker or a banner with text, was made. Detected objects that were referring to the meme text were discarded.
    \item \textbf{Multiple object detection.} In cases where the same object type appeared multiple times, the annotators would inspect the bounding boxes in the images to verify whether it is the same object. If yes, then only one occurrence was kept.
    \item \textbf{Screenshots}. Memes that are screenshots of text and do not contain visual information were disregarded from the augmentation process.
    \item \textbf{Missed objects.} The human augmentation is based on the automatically generated scene graphs and thus, no new objects or their corresponding entities were added.
    \item \textbf{Bounding boxes.} The bounding boxes that were drawn on the image during the automatic augmentation are not used in our study, hence they were not changed in the human augmentation.
    \item \textbf{Incorrect objects.} For each detected object, the evaluators determined if its type was correct or not. If not, then the correct type for this object was indicated, when it was possible. In cases where the object did not exist in the image at all, it was simply removed.
    \item \textbf{Relation correction.} For each relation, the evaluators determined if it was correct or not. If the type of objects that the relation referred to was incorrect, then, the type was replaced with the correct one. Following the same approach as with the objects, no new relations were added neither was an object that was not detected added to a relation.
    \item \textbf{Knowledge base.} Each object was mapped to a specific entity when possible by adding a link to Wikidata.
\end{itemize}

After the guidelines were finalized, the entire dataset was augmented. The inter-annotator agreement and Cohen's kappa regarding the correctness of the detected objects were 83.84\% and 0.60 respectively, while for the correctness of the detected relations it was 78.05\% and 0.53. The first evaluator found that 12.95\% of the automatically detected objects needed correction.
The second evaluator found a larger percentage of incorrect objects around 21.86\%. Similar outcomes were observed for the relations, where the first evaluator found 22.13\% of the relations to be incorrect, while the second found 31.48\%.
The number of objects found incorrect by both evaluators was 1,127 and for these cases, they agreed on the correct object 282 times. This low agreement shows that deciding about the exact type of objects shown in an image is a difficult task. The relations found incorrect by both evaluators were 1,820. Regarding the knowledge base linking, more inconsistencies were found and the evaluators added the same link in 4,314 out of 9,666 cases. In Fig. \ref{fig:annotation_examples}, we see examples of cases that caused the low agreement. Sub-figures (a) and (b) contain detected objects that both evaluators agreed are incorrect (e.g., ``10-tree'' in (a) and ``13-wheel'' in (b)). However, they did not add the same correct object. For example, the object ``13-wheel'' in (b), was corrected as ``shoe'' by one evaluator and as ``foot'' by the other. Both of these corrections can be considered valid. Regarding adding the links to Wikidata, similar uncertainties can be found. In cases of disagreement, we chose the correct object or link for the final dataset by the following heuristic: for each annotated alternative, we counted its occurrence in the part of the dataset that had 100\% agreement, i.e., the more frequent object or link was finally chosen.
For the entity links, there are cases that the evaluators added a different link, but both of the links were correct. In sub-figure (c), we see such a case in which ``0-face'' depicts both ``Bernie Sanders'' and ``Yoda''. Hence, both links were added to the final augmentations.
In cases where the evaluators disagreed regarding the correctness of an object or relation, they were removed.

To conclude our analysis based on human augmentation, we used the results of the aforementioned process as inputs to the models described in Section \ref{ssec:benchmark_models}. We experimented with two different settings for training and testing the models. First, we used the results of the manual augmentation, which are the corrected scene graphs and manually linked knowledge, both for training and inference (manual/manual). Second, we combined the two augmentations and trained on the manual ones, and tested on the automatic ones (manual/automatic). In this case, the automatic augmentations are the scene graphs corrected automatically based on the manual corrections of the training data. We kept in the development and test scene graphs only the objects that were manually marked as correct at least once in the training scene graphs and removed the rest (i.e., the ones that the evaluators found were always falsely detected by the scene graph model). The knowledge base information consisted of the descriptions of the automatically detected text entities (See Subsection \ref{ssec:linking}).
We used the same experimental setup as for the MemeGraphs method described in Subsection \ref{ssec:setup}. The average results over the 20 runs are shown in Table \ref{tab:mean_manual_results} and the score of the models performing best on the development set in Table \ref{tab:best_dev_manual_results}. We observe that in both settings MemeGraphs[SceneGr+Know] achieves the best F1 score in Table \ref{tab:best_dev_manual_results}, similar to the fully automatic setting (Table \ref{tab:best_dev_auto_results}). On the other hand, when looking at the average, the TxtBERT with only the scene graphs as input obtains the best score (MemeGraphs[SceneGr])in both the manual/manual and the manual/automatic settings.

 \begin{table}[t]
 \centering
    \resizebox{\textwidth}{!}{\begin{tabular}{l c c c c c c} \hline
    \toprule
        \multirow{2}{*}{\textbf{Model}} & \multicolumn{3}{c}{\textbf{Manual/manual}} & \multicolumn{3}{c}{\textbf{Manual/automatic}} \\ \cline{2-7} 
         & \textbf{P} & \textbf{R} & \textbf{F1} & \textbf{P} & \textbf{R} & \textbf{F1} \\ \hline
        ImgBERT & 0.394 \textpm 0.041 & 0.522 \textpm 0.080 & 0.408 \textpm 0.048 & 0.394 \textpm 0.041 & 0.522 \textpm 0.080 & 0.408 \textpm 0.048 \\
        TxtBERT & \textbf{0.457 \textpm 0.020} & 0.516 \textpm 0.063 & 0.442 \textpm 0.032 & 0.457 \textpm 0.020 & 0.516 \textpm 0.063 & 0.442 \textpm 0.032 \\ 
        MemeGraphs[SceneGr] & 0.410 \textpm 0.024 & \textbf{0.604 \textpm 0.051} & \textbf{0.474 \textpm 0.029} & 0.404 \textpm 0.024 & \textbf{0.592 \textpm 0.050} & \textbf{0.467 \textpm 0.029} \\ 
        MemeGraphs[Know] & 0.408 \textpm 0.024 & 0.507 \textpm 0.050 & 0.434 \textpm 0.030  & \textbf{0.460 \textpm 0.050} & 0.309 \textpm 0.079 & 0.254 \textpm 0.049 \\
        MemeGraphs[SceneGr+Know] & 0.436 \textpm 0.031 & 0.560 \textpm 0.051 & 0.443 \textpm 0.033 & 0.368 \textpm 0.030 & 0.331 \textpm 0.079 & 0.280 \textpm 0.045 \\ \hline
\end{tabular}}
\caption{Precision (P), recall (R) and F1 scores averaged over twenty runs and standard error of mean on the test set for each setting.}
\label{tab:mean_manual_results}
\end{table}

 \begin{table}[t]
 \centering
 \setlength{\tabcolsep}{8pt}
    \begin{tabular}{l c c c c c c} \hline
    \toprule
        \multirow{2}{*}{\textbf{Model}} & \multicolumn{3}{c}{\textbf{Manual/manual}} & \multicolumn{3}{c}{\textbf{Manual/automatic}} \\ \cline{2-7} 
        & \textbf{P} & \textbf{R} & \textbf{F1} & \textbf{P} & \textbf{R} & \textbf{F1} \\ \hline
        ImgBERT & 0.389 & \textbf{1.000} & 0.560 & 0.389 & \textbf{1.000} & 0.560 \\ \hline
        TxtBERT & 0.403 & 0.897 & 0.556 & 0.403 & 0.897 & 0.556 \\ \hline
        MemeGraphs[SceneGr] & 0.419 & 0.845 & 0.560 & 0.417 & 0.828 & 0.555 \\ \hline
        MemeGraphs[Know] & 0.400 & 0.655 & 0.497 & 0.406 & 0.931 & 0.565 \\ \hline
        MemeGraphs[SceneGr+Know] & \textbf{0.424} & 0.862 & \textbf{0.568} & \textbf{0.422} & 0.931 & \textbf{0.581} \\ \hline
\end{tabular}
\caption{Precision (P), recall (R) and F1 scores on the test set of the best model based on the development set for each setting.}
\label{tab:best_dev_manual_results}
\end{table}

\subsection{Discussion}

We observe that overall our proposed method outperforms its competitors in terms of F1 score in all settings (fully automatic, manual/manual, manual/automatic). Between the different settings, we see that the models with the fully automatic setting have the best scores (MemeGraphs[Know] in Table \ref{tab:mean_auto_results} and MemeGraphs[SceneGr+Know] in Table \ref{tab:best_dev_auto_results}), even though manual augmentations would be expected to be more accurate than the automatic ones. Furthermore, not only the best score is achieved by the MemeGraphs[Know] model in the fully automatic setting, but this model's performance is improved in this setting compared to the manual ones. The scene graphs infused model also achieves better results in the fully automatic setting, showing that the automatically produced scene graphs are accurate enough and no manual correction is needed. In the manual/automatic setting, all the models performed worse compared to the other settings. This fact shows that models trained on manual annotations are not able to generalize in the automatic setting. This holds true especially in the MemeGraphs[Know] model, since in the automatic setting no information for the type of the objects exists in the input.
To gain insights into that behavior, we analyze the different challenges that were faced in the manual and the automatic augmentation and compare their results.

During the manual augmentation, both correcting the scene graphs and adding background knowledge were found challenging. Regarding the scene graphs, memes contain complex information and images, which made the correction of objects difficult for the evaluators (see Subsection \ref{ssec:humanannotation}). Background knowledge for specific entities was also difficult to add for three main reasons: 1. the evaluators may not know the person depicted in the image, 2. many memes were screenshots of posts, so there was no actual visual information, and 3. meme texts often refer to entities that are not shown in the image. This resulted in detecting entities for only 409 memes out of 743. The automatic entity detection, on the other hand, was based on the text, which assisted in overcoming the three aforementioned challenges and extracted entities for all the memes.

Regarding the automatic augmentation based on the entities detected by the NER model, the main challenge consists of linking the detected entities to the knowledge base. Even though the model managed to extract entities from all the texts showing that we can obtain rich information, linking them to the knowledge base was not easy. Many times the entities appear in the text with only their first name, e.g., ``Hillary'', or their first and last name concatenated, e.g., ``donaldtrump'', or are detected alongside some other word, e.g., ``green Bernie''. When these entities are searched for in the knowledge base, the results might not be accurate and contain data entries for many entities from which we choose the first as the most related one. However, this can lead to adding a link to the wrong entity.

\section{Conclusion}

In order to understand memes, it is necessary to correctly interpret the image, and the text and to connect it with appropriate general background knowledge (outside of the meme).
In this work, we introduced models infused with scene graphs and world knowledge retrieved from WikiData.
As a foundational representation, scene graphs were automatically generated, which relate the most important objects in the meme image to each other. 
Typed objects from the scene graph and named entities from the text were extracted automatically and linked to WikiData.
This structured information (scene graph and information from WikiData) was then serialized as a sequence of tokens and concatenated with the original text from the meme for classification with a Transformer language model.
We found that adding the graph representation and knowledge from Wikidata improved performance on hateful meme detection compared to classification on text alone, and compared to a multimodal model based on pre-trained image embeddings in addition to text.
We also provide a dataset with human corrections of the automatically generated graphs, and an analysis that shows that the (uncorrected) automatic graphs and the corrected ones perform similarly well for hatefulness detection with our approach.

\section*{Acknowledgments}
This research was funded by the Deutsche Forschungsgemeinschaft (DFG, German Research Foundation) - RO 5127/2-1 and the Vienna Science and Technology Fund (WWTF)[10.47379/VRG19008]. We thank Christos Bintsis for participating in the manual augmentation. We also thank Matthias Aßenmacher and the anonymous reviewers for their valuable feedback.

\bibliographystyle{splncs04}  
\bibliography{references}

\end{document}